\documentclass[10pt,twocolumn,letterpaper]{article}

\usepackage{cvpr}
\usepackage{times}
\usepackage{epsfig}
\usepackage{graphicx}
\usepackage{amsmath}
\usepackage{amssymb}
\usepackage{color}
\usepackage{booktabs,threeparttable}
\usepackage{colortbl}
\usepackage[export]{adjustbox}
\usepackage{caption}
\usepackage{multirow}

\usepackage{bbm}

\usepackage[pagebackref=true,breaklinks=true,letterpaper=true,colorlinks,bookmarks=false]{hyperref}

\cvprfinalcopy 

\def\cvprPaperID{} 

\begin{document}

\title{MSE-Nets: Multi-annotated Semi-supervised Ensemble Networks for Improving Segmentation of Medical Image with Ambiguous Boundaries}

\author{Shuai Wang$^\star$\\
Hangzhou Dianzi University, China\\
{\tt\small shuaiwang.tai@gmail.com}
\and
Tengjin Weng$^\star$\\
Zhejiang Sci-Tech University, China\\
{\tt\small wtjdsb@gmail.com}
\and
Jingyi Wang\\
Shandong University, China\\
{\tt\small jywang@mail.sdu.edu.cn}
\and
Yang Shen\\
Lishui University, China\\
{\tt\small tlsheny@163.com}
\and
Zhidong Zhao\\
Hangzhou Dianzi University \\ Center, Dallas, TX, USA\\
{\tt\small zhaozd@hdu.edu.cn}
\and
Yixiu Liu\\
Hangzhou Dianzi University, China\\
{\tt\small liuyixiu@hdu.edu.cn}
\and
Pengfei Jiao\\
Hangzhou Dianzi University, China\\
{\tt\small pjiao@hdu.edu.cn}
\and
Zhiming Cheng\\
Hangzhou Dianzi University, China\\
{\tt\small chengzhiming1118@gmail.com}
\and
Yaqi Wang\\
Communication University of Zhejiang, China\\
{\tt\small wangyaqi@cuz.edu.com}
}

\maketitle

\begin{abstract}
Medical image segmentation annotations exhibit variations among experts due to the ambiguous boundaries of segmented objects and backgrounds in medical images. Although using multiple annotations for each image in the fully-supervised has been extensively studied for training deep models, obtaining a large amount of multi-annotated data is challenging due to the substantial time and manpower costs required for segmentation annotations, resulting in most images lacking any annotations. 
To address this, we propose Multi-annotated Semi-supervised Ensemble Networks (MSE-Nets) for learning segmentation from limited multi-annotated and abundant unannotated data. 
Specifically, we introduce the Network Pairwise Consistency Enhancement (NPCE) module and Multi-Network Pseudo Supervised (MNPS) module to enhance MSE-Nets for the segmentation task by considering two major factors:
(1) to optimize the utilization of all accessible multi-annotated data, the NPCE separates (dis)agreement annotations of multi-annotated data at the pixel level and handles agreement and disagreement annotations in different ways,
(2) to mitigate the introduction of imprecise pseudo-labels, the MNPS extends the training data by leveraging consistent pseudo-labels from unannotated data.
Finally, we improve confidence calibration by averaging the predictions of base networks.
Experiments on the ISIC dataset show that we reduced the demand for multi-annotated data by 97.75\% and narrowed the gap with the best fully-supervised baseline to just a Jaccard index of 4\%.
Furthermore, compared to other semi-supervised methods that rely only on a single annotation or a combined fusion approach, the comprehensive experimental results on ISIC and RIGA datasets demonstrate the superior performance of our proposed method in medical image segmentation with ambiguous boundaries.
\end{abstract}


\section{Introduction}

Medical image segmentation plays a critical role in computer-aided diagnosis systems, enabling precise delineation of structures and regions of interest. Convolutional neural networks (CNNs) have emerged as powerful tools for automatic segmentation tasks, exhibiting impressive performance in various medical imaging modalities. These networks leverage their ability to learn complex patterns and features from large amounts of annotated data to segment medical images accurately. 
However, collecting ground truth annotations for semantic segmentation is considerably more expensive than for other visual tasks such as classification and object detection due to the dense annotations involved. 
While this can be partly mitigated by outsourcing the annotation process to non-experts, the presence of multiple object classes in a scene, coupled with factors like illumination, shading, and occlusion, makes delineating precise object boundaries an ambiguous and laborious task, resulting in some unavoidable noise in the annotations.
Despite significant research efforts to develop noise-resistant segmentation networks \cite{shi2021distilling, weng2023learning, han2018co,zhang2020robust,zhu2019pick, xu2022anti,9410359}, it remains challenging to eliminate deep-rooted biases present in annotations \cite{vorontsov2021label}.
A widely adopted strategy for addressing imprecision annotations is the utilization of multi-annotated.
It is worth noting that using multiple annotations\cite{almazroa2017agreement, orlando2020refuge, armato2011lung,zhang2023multi,nguyen2019deepusps, wang2022multi,9508607} for each image has been widely studied for training deep models.

\begin{figure}[!t]
\centerline{\includegraphics[width=\linewidth]{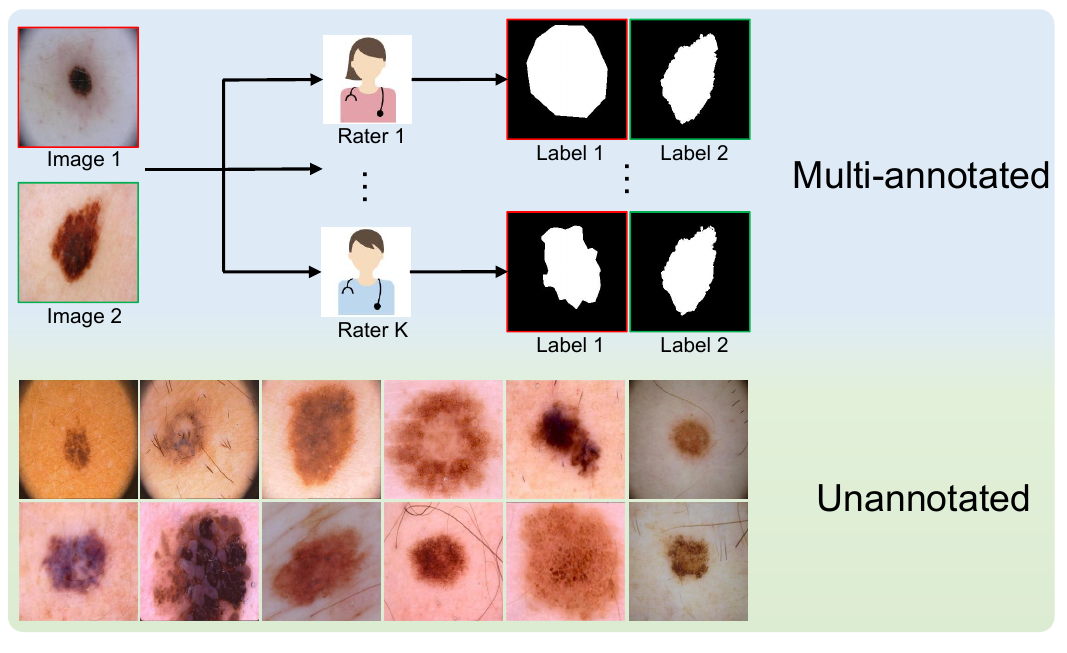}}
    \caption{Data collection of our proposed method, which contains a small amount of multi-annotated data and a large amount of unannotated data.}
    \label{fig1}
\end{figure}

There are methods that focus on finding efficient and reasonable fusion strategies. These strategies aim to combine multiple annotations to obtain a more reliable and accurate segmentation. Examples of such fusion methods include STAPLE~\cite{warfield2004simultaneous} and majority voting, where the consensus among annotations is used to generate a final annotation. 
However, manual annotation of anatomical regions of interest can be very subjective, and there can be considerable disagreement among annotators and within annotators even among experts in multiple medical imaging modalities, making it difficult to obtain a centralized gold standard annotation for model training and evaluation.
Therefore, some researchers have proposed label selection strategies, where a carefully selected subset of images is used to train the segmentation model, and label sampling strategies, where labels are randomly drawn from a multi-annotator label bank at each training iteration to generate multiple predictions at different sensitivity settings to prevent overconfidence of a single network.
There has been extensive research on training deep models using multiple annotations per image under full supervision. However, an unavoidable challenge is that most of the data lacks any annotations, given the huge costs involved (both in terms of labor and time).

When we convert the application scenario to a situation where there is only a small amount of multi-annotated data and a large amount of unannotated data, neither the careful label subset selection strategy nor the label sampling strategy is applicable. In the case of insufficient multi-annotated data, it is necessary to make full use of each expert's annotations and it is difficult to fully assess the differences between different experts. Moreover, how to utilize unannotated data is also a huge challenge. While exploring available unlabeled images is indeed valuable for training segmentation models, it is important to note that the application of semi-supervised semantic segmentation has primarily focused on scenarios with well-defined and non-ambiguous boundaries~\cite{french2019semi,kim2020structured,ouali2020semi,ke2019dual}. When it comes to the specific challenge of handling ambiguous boundaries in semi-supervised segmentation, the exploration and development of dedicated techniques are still relatively limited. Ambiguous boundaries introduce additional complexities, as labeled data cannot provide accurate prior knowledge. The inaccurate information learned from labeled data can transfer to unlabeled data, making it challenging to enforce consistency among different boundary interpretations. Moreover, while consistency regularization can effectively leverage unlabeled data to enhance the segmentation model's performance and reduce the reliance on labeled data, it may not fully address the inherent uncertainties and subjective interpretations associated with ambiguous boundaries.

To tackle these issues, we introduce the Multi-annotated Semi-supervised Ensemble Networks (MSE-Nets) with the backbone network consisting of multiple LinkNets~\cite{chaurasia2017linknet} initialized differently and designed for segmentation from a limited multi-annotated dataset and an extensive unannotated dataset.
Fig.~\ref{fig1} illustrates the data collection process for our approach, involving two main components: (1) a small multi-annotated dataset curated by $K$ experts ($K \ge 2$) and (2) a significantly larger unannotated dataset.
For these distinct data, we propose the Network Pairwise Consistency Enhancement (NPCE) and Multi-Network Pseudo Supervised (MNPS) modules, serving two primary purposes: (1) maximizing the utilization of all available multi-annotated data and (2) mitigating the impact of imprecise pseudo-labels from the unannotated dataset on the network.
\begin{itemize}
\item[$\bullet$] We combine multi-annotated and semi-supervised segmentation and propose the MSE-Nets, aiming to improve the performance of ambiguous boundaries medical image segmentation in scenarios with a small amount of multi-annotated data and a large number of unannotated.
\item[$\bullet$] We propose the NPCE module for separating pixel-level (dis)agreement information from multi-annotated data for two purposes: (1) agreement information is directly input into the network as reliable prior knowledge and (2) disagreement information is replaced based on whether the prediction results are consistent for label refinement.
\item[$\bullet$] We propose the MNPS module use the predicted consistent masks of multiple networks as the ground truth for unannotated images. The MNPS serves two benefits: (1) strengthening the consistency between networks by incorporating additional intrinsic image knowledge from a substantial volume of unannotated data, which can be transferred to enhance the prediction consistency of multi-annotated data, and (2) preemptively circumventing the adverse impact of imprecise pseudo-labels on the network's learning.
\end{itemize}

\begin{figure}[!t]
\centerline{\includegraphics[width=\linewidth]{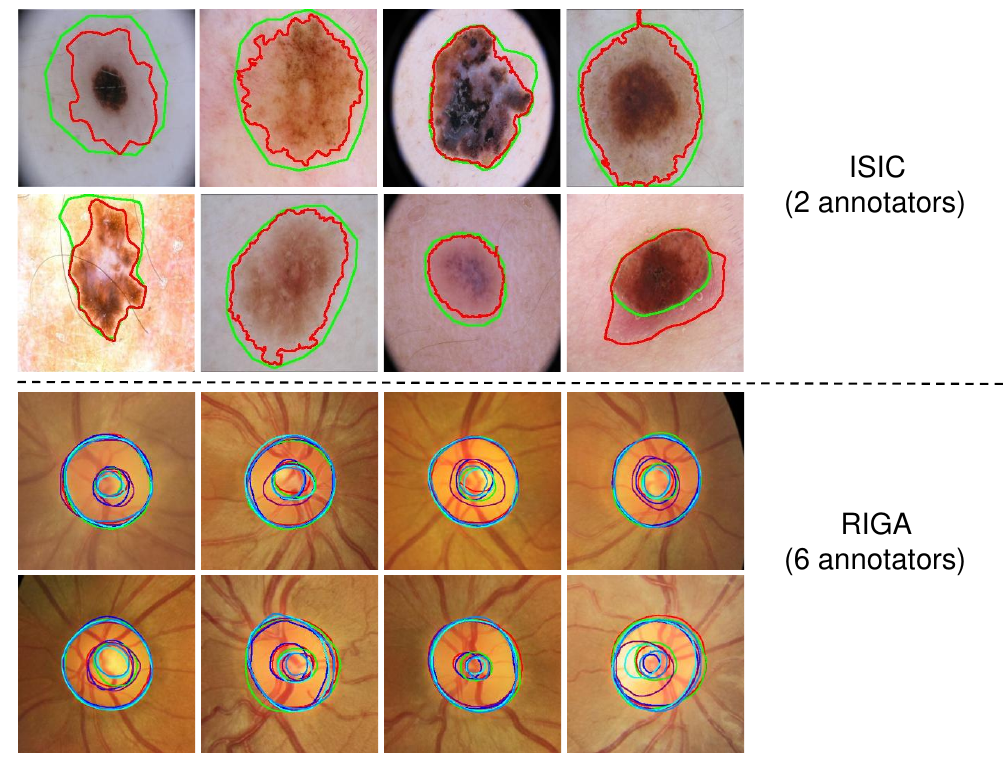}}
    \caption{Image of the skin lesion samples from the ISIC archive and the optic disc and cup segmentation samples from RIGA with multiple boundary annotations from different annotators (in different colors).}
    \label{fig2}
\end{figure}

\section{Related Work}
\subsection{Medical Image Segmentation}

Accurate segmentation of internal structures from medical images is paramount for various clinical applications. Medical image segmentation refers to the process of separating different tissues, organs, or pathological regions in the image for further analysis and diagnosis. For instance, accurate segmentation plays a crucial role in tumor detection and localization, lesion analysis, surgical planning, and navigation, among others.
However, medical images often exhibit complex anatomical structures and variabilities, making the segmentation task complex and challenging. For example, boundaries between tumors and organs may be ambiguous, tissues can vary significantly in shape and size, and images may contain noise and artifacts. These factors make it difficult for traditional image processing approaches to achieve satisfactory results.

To address these challenges, deep learning methods, particularly those based on CNNs, have emerged as the primary tools for medical image segmentation.
Many CNN-based methods~\cite{long2015fully,badrinarayanan2017segnet,ronneberger2015u,chen2017deeplab} have been developed for performing segmentation tasks. 
However, delineating precise object boundaries is a fuzzy and laborious task due to the involvement of dense annotations, leading to disagreements among annotators. The existence of multiple annotations further poses a challenge in determining the ideal ground truth for assessing model performance.

\subsection{Multi-annotated Medical Image Segmentation}

Annotations for medical image segmentation with ambiguous boundaries, even when performed by experts, are inevitably contained by noise and bias.
Fig.~\ref{fig2} visually illustrates the differences in annotations of ambiguous boundaries between different annotators. A widely adopted strategy for addressing medical image segmentation with ambiguous boundaries is the utilization of multi-annotated. Some existing multi-annotated methods have demonstrated their superior performance compared to using a single annotation. 

Ribeiro~\textit{et al.}~\cite{ribeiro2019handling} introduced an approach that enhances the agreement between annotators by utilizing morphological image processing operations, such as opening and closing, convex hulls, and bounding boxes, to eliminate annotator-specific details from the segmentation masks. By applying these operations, they aimed to condition the segmentation masks and interpret this process as a denoising procedure that removes annotator-specific variations. The same authors suggested a strategy for training their segmentation model using a carefully selected subset of images. Specifically, they excluded samples with an average pairwise Cohen's kappa score below 0.5, ensuring that only segmentation annotations with significant agreement between annotators are used to train the model \cite{ribeiro2020less}.

Moreover,  Zhang~\textit{et al.}~\cite{zhang2020learning} propose a neural network architecture that simultaneously learns the reliability of individual annotators and the true distribution of segmentation labels. By emphasizing the disjoint features of annotators and the true segmentation labels, the proposed framework enables effective learning of both aspects, leading to improved accuracy in estimating the underlying segmentation label distribution. Mirikharaji~\textit{et al.}~\cite{mirikharaji2021d} propose an ensemble approach based on FCNs~\cite{long2015fully} for segmentation tasks. The primary focus of their method is to handle contradictory annotations present in the training data, which result from disagreements between annotators. Additionally, their approach incorporates improved confidence calibration predictions from the underlying model. The ensemble framework effectively addresses the challenge of contradictory annotations and enhances the overall segmentation performance. 
Ji~\textit{et al.}~\cite{ji2021learning} introduced MRNet, a method that leverages the professional expertise of each rater as prior knowledge to generate high-level semantic features. The proposed approach also involves reconstructing multi-rater ranks based on initial predictions and exploiting the (in-)consistent cues from multiple raters to enhance segmentation performance.

However, constructing large-scale multi-annotated datasets to train CNN-based methods for medical image segmentation faces great challenges. The process is not only resource-intensive but also demands extensive domain expertise. As a result, assembling comprehensive multi-annotated datasets becomes a time-consuming and sometimes impractical endeavor. When the volume of multi-annotated data sharply decreases, it becomes necessary to fully utilize each expert's annotations, rendering label selection-based methods impractical in practice. Meanwhile, the differences between annotations will be reduced, making it difficult for the label sampling strategy to produce differentiation and to fully assess the professionalism between different annotations.

\subsection{Semi-supervised Medical Image Segmentation}
Semi-Supervised Learning (SSL) method can effectively extract informative features from unlabeled data to potentially alleviate the limitation brought by limited labeled data.
Many efforts have been made in semi-supervised medical image segmentation. Consistency regularization is widely studied for semi-supervised segmentation. Mean-Teacher (MT) ~\cite{tarvainen2017mean} is a classic SSL framework based on consistency regularization. Meanwhile, many works extend MT in different ways to build the SSL framework. UAMT~\cite{yu2019uncertainty} utilizes uncertainty information to guide the student network to learn gradually from reliable and meaningful targets provided by the teacher network. SASSNet~\cite{li2020shape} utilizes unlabeled data to enforce geometric shape constraints on segmentation results. DTC~\cite{luo2021semi} proposes a dual-task consistency framework by explicitly building task-level regularization. 
Other methods, such as CPS~\cite{chen2021semi}, utilize two networks with the same structure but different initializations, imposing constraints to ensure their outputs for the same sample are similar. ICT~\cite{verma2022interpolation} encourages the coherence between the prediction at an interpolation of unlabeled points and the interpolation of the predictions at those points. These SSL methods further improve the effectiveness of semi-supervised medical image segmentation.
BCP~\cite{bai2023bidirectional} introduces a bidirectional CutMix~\cite{yun2019cutmix} approach to facilitate comprehensive learning of common semantics from labeled and unlabeled data in both inward and outward directions.

However, the exploration and development of dedicated techniques for addressing the specific challenge of handling ambiguous boundaries in semi-supervised segmentation are still relatively limited. Ambiguous boundaries introduce additional complexities, as labeled data cannot provide accurate prior knowledge for these regions. Inaccurate information learned from labeled data may be transferred to unlabeled data, making it challenging to achieve consistency between different boundary interpretations.
 
\begin{figure*}[!t]
\centerline{\includegraphics[width=\linewidth]{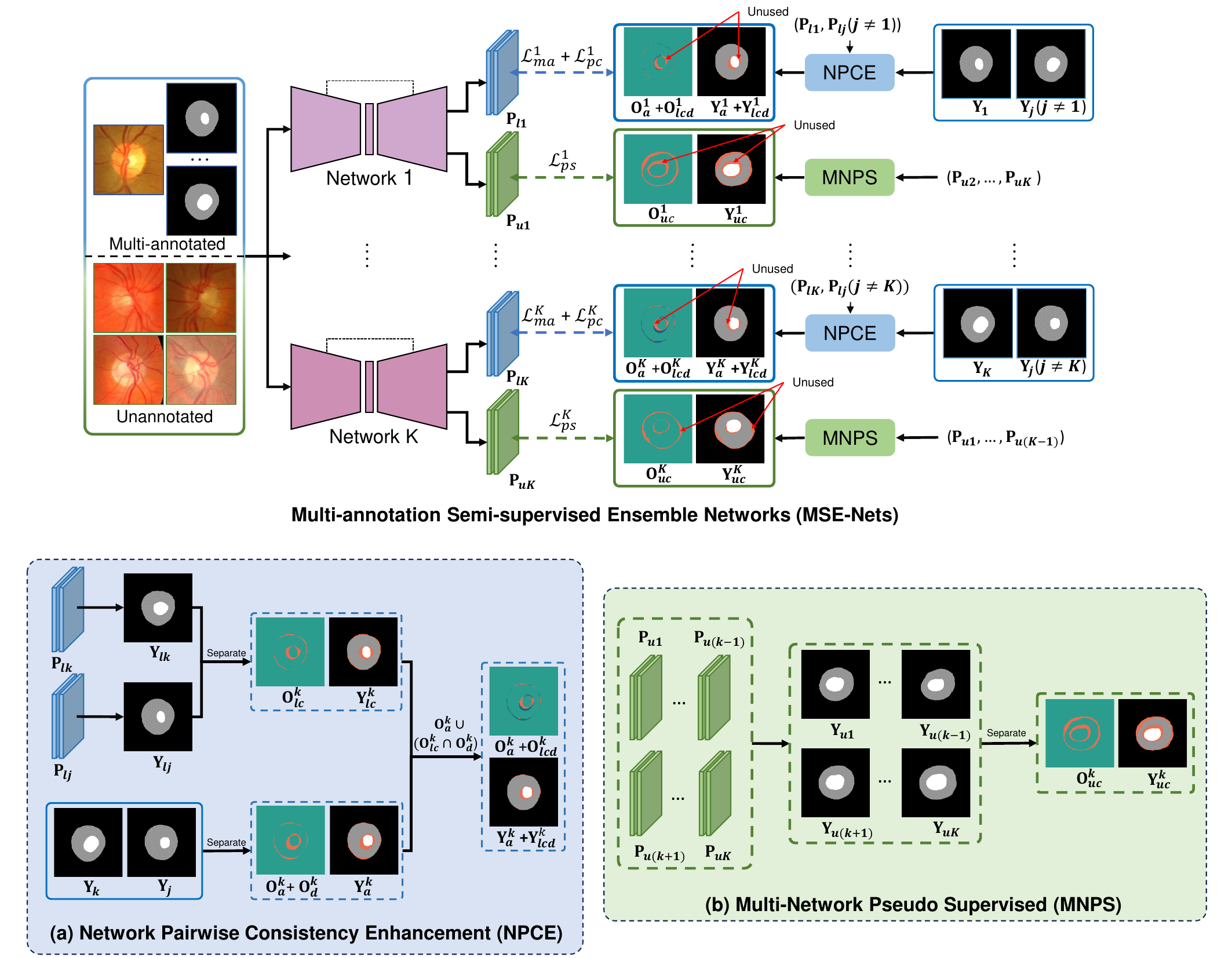}}
    \caption{Illustration of our proposed method. The upper part is the overall network architecture, and the lower part is the two modules NPCE and MNPS. Our method is constructed by $K$ networks corresponding to $K$ annotations, where each network incorporates the NPCE module and MNPS module during every iteration. (a) The NPCE utilizes pixel-level information separation to fully utilize precise annotation information, which includes two aspects: (1) agreement annotation information as basic reliable prior knowledge; and (2) partially disagreement information between annotations is compensated with consistent information between network predictions.
    (b) The MNPS avoids imprecise information in network learning by using consistent pseudo-labels between networks as a reliable ground for unannotated images.}
    \label{fig3}
\end{figure*}

\begin{figure}[!t]
\centerline{\includegraphics[width=\linewidth]{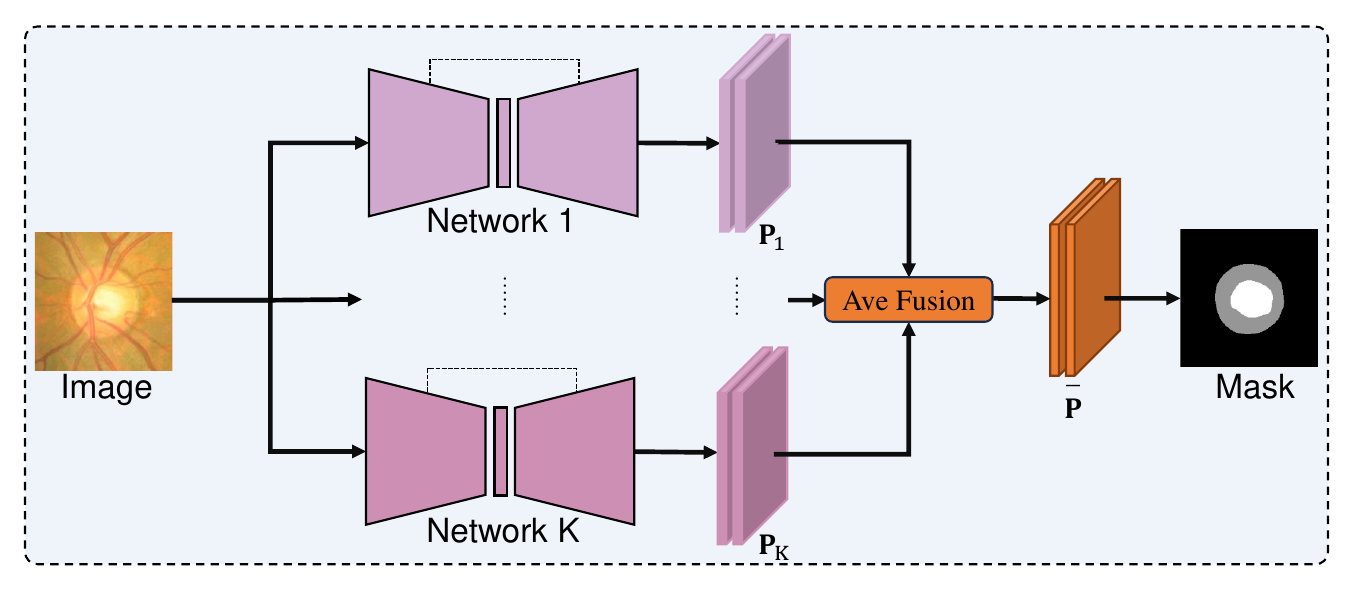}}
    \caption{Illustration the stage of inference of MSE-Nets, the predicted probability maps from the $K$ networks are averaged fusion to obtain the final prediction mask.}
    \label{fig4}
\end{figure}

\section{Methodology}
\subsection{Framework}
In this work, we proposed a novel framework MSE-Nets for learning segmentation from a small amount of multi-annotated data and a large amount of unannotated data. To simplify the description of our methodology, we define $N$ samples to represent the total data and $M$ samples to represent the multi-annotated data, while the remaining $N-M$ samples represent the unannotated data. We denote multi-annotated data as $\mathbf{D}_m=\{\mathbf X_{(i)},  \mathbf Y_{(i)}^{1}, \mathbf Y_{(i)}^{2},...,\mathbf Y_{(i)}^{K}\}^M_{i=1}$, where $ \mathbf X$ represent the input image and $\mathbf Y^{k}$ represent the ground truth mask of $k^{th}$ annotators, here $K$ is the number of annotators. The unannotated data is represented as $\mathbf{D}_u=\{\mathbf X_{(i)}\}^N_{i=M+1}$. The proposed MSE-Nets is trained by the combined dataset $\{\mathbf{D}_m, \mathbf{D}_u\} $.

Fig.~\ref{fig3} illustrates the method of our proposed. The MSE-Nets is constructed by $K$ networks corresponding to $K$ annotations, where each network incorporates the NPCE module and MNPS module during every iteration. The NPCE uses reliable information to constrain network learning and ensure consistency between networks by comparing prediction information and corresponding annotation information between networks.
The MNPS uses multi-network predicted consistent pseudo-labels as reliable ground truth for unannotated images to extend the training set.
The proposed approach offers several benefits:
(i) Removing disagreement boundaries annotations: Excluding disagreement annotations from the training data helps the network to focus on learning accurate annotation knowledge.
(ii) Refining the annotations of disagreement boundaries: Replacing disagreement pixel labels with pixel labels with consistent predictions between the network to further provide reliable information.
(iii) Introducing additional image-only knowledge: Pseudo-labels derived from consistent predictions on large amounts of unannotated data extend the training set, and this knowledge can be transferred into multi-annotated data to further improve network performance.
At the inference stage, the predicted probability maps from the $K$ networks are averaged fusion to obtain the final prediction mask. Fig.~\ref{fig4} illustrates the inference process of our proposed method. 
More details of MSE-Nets will be described in the following sections.
\subsection{Network Pairwise Consistency Enhancement (NPCE)}
\subsubsection{Separating (Dis-)Agreement Annotations of Multi-annotated Data}
Annotation errors, which can arise from inter-annotator differences and ambiguous boundaries, have the potential to significantly impact network performance, often leading to suboptimal results and reduced generalization capabilities.
Therefore, we consider separating reliable and unreliable annotations at the pixel level based on different annotations from multiple experts before network training starts. 

Formally, give an image $\mathbf{X} \in \mathbf{D}_m$, the multi-annotated ground truth masks are represented $\mathbf{Y} = \{\mathbf{Y}_1, \mathbf{Y}_2,...,\mathbf{Y}_K\}$, where $\mathbf{X} = \{X_i\}_{i=1}^n$ and $n = w \times h$ is the number of pixels. The ground truth mask of the $k^{th}$ annotator is expressed as $\mathbf{Y}_k = \{Y_k^i\}_{i=1}^n$ and $Y_k \in \{1,2,..., C\}$, where $C$ is the number of semantic classes. Our goal is to train ensemble networks containing $K$ networks, with each network corresponding to one annotator.
For each network, another network is randomly selected as the comparison network. Assume that the current network is the $k^{th}$ network and the corresponding annotated is $\mathbf{Y}_k$, and the randomly selected comparison network is the $j^{th}$ network and the corresponding annotated is $\mathbf{Y}_j$. 

In order to obtain agreement information from $\mathbf{Y}_k$ and $\mathbf{Y}_j$, we define the agreement pixels set $\mathbf{O}_a^k$ and the disagreement pixels set $\mathbf{O}_d^k$, which are calculated by the following formula :
\begin{equation}
\begin{split}
&\mathbf{O}_a^k = \{i \mid Y_k^i = Y_j^i\}_{i=0}^{n}, \\
&\mathbf{O}_d^k = \{i\mid  Y_k^i \neq Y_j^i\}_{i=0}^{n}.
\end{split}
\end{equation}
The corresponding label $\mathbf{Y}_a^k$ of $\mathbf{O}_a^k$ is expressed as:
\begin{equation}
\begin{split}
&\mathbf{Y}_a^k = \{Y_k^i \mid Y_k^i = Y_j^i\}_{i=0}^{n}.
\end{split}
\end{equation}

The $s_a^k = |\mathbf{O}_a^k|$ and $s_d^k = |\mathbf{O}_d^k|$ represent the number of pixels contained in the $\mathbf{O}_a^k$ and $\mathbf{O}_d^k$ respectively.


The (dis-)agreement pixels set will constrain network learning in different ways, which will be introduced in subsequent sections respectively.

\subsubsection{Learning Reliable Knowledge from Agreement Pixels}
Annotation results may contain noise due to inter-annotator differences and ambiguous boundaries. However, when multiple annotators have agreement insights on the same pixel, the consensus among annotators provides strong evidence that the pixel label is considered a genuine classification rather than noise. 

We input $\mathbf{X} \in \mathbf{D}_m$ into the $k^{th}$ and $j^{th}$ networks, and get the corresponding predicted probability maps as $\mathbf{P}_{lk}$ and $\mathbf{P}_{lj}$. The corresponding predicted masks $\mathbf{Y}_{lk} = \{Y_{lk}^i\}_{i=0}^n$ and $\mathbf{Y}_{lj} = \{Y_{lj}^i\}_{i=0}^n$ are calculated by:
\begin{equation}
\begin{split}
& \mathbf{Y}_{lk} = \mathop{\arg\max}\limits_{c} \mathbf{P}_{lk}(c, n),\\
& \mathbf{Y}_{lj} = \mathop{\arg\max}\limits_{c} \mathbf{P}_{lj}(c, n).
\end{split}
\end{equation}

We directly perform Cross-Entropy (CE) loss on these agreement pixels. Therefore, the multi-annotated agreement loss $\mathcal{L}_{ma}$ for the $k^{th}$ network is denoted as:
\begin{equation}
\begin{split}
\mathcal{L}_{ma}^{k} =\frac{1}{s_a^k}\sum_{i=0}^{s_a^k}{\ell_{ce}(\mathbf{P}_{lk}^{\mathbf{O}_a^k[i]},\mathbf{Y}_a^k[i])}.
\end{split}
\end{equation}
\subsubsection{Refining Disagreement Pixels for Further Improvement}
The agreement pixels set provides accurate prior information for the network, but the disagreement pixels set cannot be discarded. 
The networks with different initializations produce consistent predictions for the same pixel, it can be considered that the predictions at this pixel location are relatively reliable. This consistency indicates that the network has learned similar features or weights under different initialization conditions, making the predictions for this pixel relatively stable.
However, conversely, when the networks with different initializations produce inconsistent predictions for the same pixel, it implies that the predictions at this pixel location are not reliable. This inconsistency indicates that the network's behavior is sensitive to initialization conditions, making the predictions for this pixel less stable.

We define the prediction consistency pixels set as $\mathbf{O}_{lc}^{k}$ and the corresponding label as $\mathbf{Y}_{lc}^{k}$, which are calculated by $\mathbf{Y}_{lk}$ and $\mathbf{Y}_{lj}$ :
\begin{equation}
\begin{split}
&\mathbf{O}_{lc}^{k} = \{i \mid Y_{lk}^i = Y_{lj}^i\}_{i=0}^{n},\\
&\mathbf{Y}_{lc}^{k} = \{Y_{lk}^i \mid Y_{lk}^i = Y_{lj}^i\}_{i=0}^{n}.
\end{split}
\end{equation}

To perform prediction consistency processing only on disagreement pixels set, we take the intersection of $\mathbf{O}_{lc}^k$ and $\mathbf{O}_d^k$ to get pixels set with prediction consistency:
\begin{equation}
\begin{split}
&\mathbf{O}_{lcd}^k = \mathbf{O}_{lc}^k \cap \mathbf{O}_d^k.
\end{split}
\end{equation}

The corresponding label of $\mathbf{O}_{lcd}^k$ is expressed as $\mathbf{Y}_{lcd}^{k}$ and the $s_{lcd}^k = |\mathbf{O}_{lcd}^k|$ represents the number of pixels contained in the $\mathbf{O}_{lcd}^k$. The $\mathbf{O}_{lcd}^k$ indicates the prediction consistency information of the $k^{th}$ networks and $j^{th}$ networks. 

We use CE loss on pixels that are considered correct between these networks but are disagreement among experts. Therefore, the prediction consistency loss $\mathcal{L}_{pc}$ for the $k^{th}$ network is denoted as:
\begin{equation}
\begin{split}
\mathcal{L}_{pc}^{k} =\frac{1}{s_{lcd}^k}\sum_{i=0}^{s_{lcd}^k}{\ell_{ce}(\mathbf{P}_{lk}^{\mathbf{O}_{lcd}^k[i]},\mathbf{Y}_{lcd}^k[i])}. 
\end{split}
\end{equation}

By repeatedly comparing the decisions of the two networks, we encourage the exchange and fusion of information.
The exchange of information helps predictions between networks gradually converge, thereby reducing potential inconsistencies and improving the accuracy of the entire network ensemble.
\subsection{Multi-Network Pseudo Supervised (MNPS)}
Exploring the availability of a substantial amount of unannotated data to further improve network performance is also of utmost importance. For predictions of the same input image, we encourage the $k^{th}$ network to maintain a high degree of similarity to the other network.
 
Similarly, we input $\mathbf{X} \in \mathbf{D}_u$ into the $k^{th}$ and other networks, and get the corresponding predicted probability maps as $\mathbf{P}_{uk}$ and $\{\mathbf{P}_{u1},...,\mathbf{P}_{u(k-1)},\mathbf{P}_{u(k+1)},...,\mathbf{P}_{uK} \}$. 
The corresponding predicted masks $\{\mathbf{Y}_{u1},...,\mathbf{Y}_{u(k-1)},\mathbf{Y}_{u(k+1)},...,\mathbf{Y}_{uK} \}$ of other networks are calculated by:
\begin{equation}
\begin{split}
& \mathbf{Y}_{uz} = \mathop{\arg\max}\limits_{c} \mathbf{P}_{uz}(c, n),
\end{split}
\end{equation}
where $z \in [1, K]$, $z \ne k$ and $\mathbf{Y}_{uz} = \{Y_{uz}^i\}_{i=0}^n$.

In order to maintain the consistency of prediction results for unannotated data across networks. It is ensured that the output result of $k^{th}$ network is similar to the predicted mask of other networks for the same unannotated image.

Therefore, we obtain all the predicted consistent pixel sets $\mathbf{O}_{uc}^{k}$ of other networks as pseudo-supervised signal for the $k^{th}$ network, which can be calculated by:
\begin{equation}
\mathbf{O}_{uc}^{k} = \{i \mid {Y}_{u1}^i=...={Y}_{u(k-1)}^i={Y}_{u(k+1)}^i=...={Y}_{uK}^i\}_{i=0}^{n}.
\end{equation}
The corresponding label $\mathbf{Y}_{uc}^{k}$ is expressed as:
\begin{equation}
\mathbf{Y}_{uc}^{k} = \{{Y}_{u1}^i \mid {Y}_{u1}^i=...={Y}_{u(k-1)}^i={Y}_{u(k+1)}^i=...={Y}_{uK}^i\}_{i=0}^{n},
\end{equation}
and $s_{uc}^k = |\mathbf{O}_{uc}^{k}|$ represents the number of pixels contained in the $\mathbf{O}_{uc}^k$.

We calculate the CE loss between the predicted probability map of the $k^{th}$ network and the predicted consistent mask of the other network to enhance the prediction consistency between networks. The pseudo-supervised loss $\mathcal{L}_{ps}$ of unannotated data for the $k^{th}$ network is denoted as:
\begin{equation}
\begin{split}
\mathcal{L}_{ps}^{k} =\frac{1}{s_{uc}^k}\sum_{i=0}^{s_{uc}^k}{\ell_{ce}(\mathbf{P}_{uk}^{\mathbf{O}_{uc}^{k}[i]},\mathbf{Y}_{uc}^k[i])}.
\end{split}
\end{equation}
 
The pseudo-supervised consistency extends the training data by utilizing consistent pseudo-labels from unannotated data and facilitates the transfer of knowledge acquired from unannotated data to annotated data, thereby further enhancing the consistency among the network ensemble and further enhancing performance.

\begin{table*}[!t]
\centering
    \renewcommand\arraystretch{1.5}
    \caption{Comparing the Segmentation Performance Based on the Jaccard Index Reported in Percent (\% ± Standard Error) on Varying Amounts of Multi-Annotated Data. The Semi-Supervised Approach Introduces Additional Unannotated Data. The Best Results are in Bold.}
    \label{tab1}
    \large
\scalebox{0.9}{
\begin{tabular}{c|cccc}
\toprule[1pt] \hline
Methods                        & \multicolumn{1}{c|}{Annotator(s)} & \multicolumn{1}{c|}{$\mathbf{D}_m$ = 30}              & \multicolumn{1}{c|}{$\mathbf{D}_m$ = 50}              & $\mathbf{D}_m$ = 70 \\ \hline \hline
\multirow{2}{*}{LinkNet~\cite{chaurasia2017linknet} (VCIP 2017)} & \multicolumn{1}{c|}{1}          & \multicolumn{1}{c|}{63.02 ± 0.37}          & \multicolumn{1}{c|}{63.49 ± 1.48}          & 63.23 ± 1.96       \\  
                               & \multicolumn{1}{c|}{2}          & \multicolumn{1}{c|}{61.62 ± 1.80}          & \multicolumn{1}{c|}{62.66 ± 2.18}          & 62.53 ± 1.12       \\ \hline \hline
\multirow{2}{*}{MT\cite{tarvainen2017mean} (NIPS 2017)}      &\multicolumn{1}{c|}{1}          & \multicolumn{1}{c|}{63.10 ± 2.19}          & \multicolumn{1}{c|}{63.97 ± 1.55}          & 64.66 ± 1.01       \\  
                               & \multicolumn{1}{c|}{2}          & \multicolumn{1}{c|}{63.01 ± 0.95}          & \multicolumn{1}{c|}{62.92 ± 1.56}          & 62.91 ± 1.53       \\ \hline
\multirow{2}{*}{UAMT\cite{yu2019uncertainty} (MICCAI 2019)}    & \multicolumn{1}{c|}{1}          & \multicolumn{1}{c|}{63.92 ± 1.34}          & \multicolumn{1}{c|}{64.60 ± 1.10}          & 63.14 ± 1.14      \\  
                               & \multicolumn{1}{c|}{2}          & \multicolumn{1}{c|}{62.86 ± 1.07}          & \multicolumn{1}{c|}{63.73 ± 0.79}          &  63.33 ± 1.52       \\ \hline
\multirow{2}{*}{CPS\cite{chen2021semi} (CVPR 2021)}     & \multicolumn{1}{c|}{1}          & \multicolumn{1}{c|}{63.10 ± 1.50}          & \multicolumn{1}{c|}{64.20 ± 0.59}          & 63.81 ± 3.21        \\  
                               & \multicolumn{1}{c|}{2}          & \multicolumn{1}{c|}{62.70 ± 1.21}          & \multicolumn{1}{c|}{62.65 ± 0.67}          &  62.64 ± 2.71      \\ \hline
\multirow{2}{*}{ICT\cite{verma2022interpolation} (Neural Networks 2022)}     & \multicolumn{1}{c|}{1}          & \multicolumn{1}{c|}{64.32 ± 1.02}          & \multicolumn{1}{c|}{64.12 ± 0.78}          & 64.40 ± 1.23       \\  
                               & \multicolumn{1}{c|}{2}          & \multicolumn{1}{c|}{63.72 ± 1.05}          & \multicolumn{1}{c|}{63.16 ± 2.52}          & 63.30 ± 0.72       \\ \hline
\multirow{2}{*}{BCP\cite{bai2023bidirectional} (CVPR 2023)}     & \multicolumn{1}{c|}{1}          & \multicolumn{1}{c|}{64.51 ± 0.92}          & \multicolumn{1}{c|}{64.66 ± 1.41}          & 63.64 ± 1.63       \\  
                               & \multicolumn{1}{c|}{2}          & \multicolumn{1}{c|}{63.09 ± 0.82}          & \multicolumn{1}{c|}{64.18 ± 0.85}          & 63.27 ± 1.24       \\ \hline \hline
\textbf{MSE-Nets (Ours)}                 & \multicolumn{1}{c|}{\textbf{1,2}}        & \multicolumn{1}{c|}{\textbf{67.34 ± 0.27}} & \multicolumn{1}{c|}{\textbf{68.27 ± 0.65}} & \textbf{68.40 ± 0.35}       \\ \hline \hline
LIS~\cite{ribeiro2020less} ($\mathbf{D}_m$ = 2333)               & \multicolumn{4}{c}{69.20}                                                                                                      \\ \hline
D-LEMA~\cite{mirikharaji2021d} ($\mathbf{D}_m$ = 2333)             & \multicolumn{4}{c}{72.11 ± 0.51}                                                                                                \\ \hline \bottomrule[1pt]
\end{tabular}}
\end{table*}

\begin{table}[!t]
\centering
    \renewcommand\arraystretch{1.5}
    \caption{Ablation Study of Using Average Fusion for MSE-Nets Inference. The Best Results are in Bold.}
    \label{tab2}
    \large
\scalebox{0.78}{
\begin{tabular}{c|c|c|c|c}
\toprule[1pt] \hline
Methods                         & Network(s) & $\mathbf{D}_m$ = 30              & $\mathbf{D}_m$ = 50              & $\mathbf{D}_m$ = 70 \\ \hline \hline
\multirow{3}{*}{MSE-Nets} & 1          & 66.49±0.46          & 67.07±0.73          & 67.52±0.33       \\
                                & 2          & 66.96±0.20          & 67.36±1.09          & 67.93±0.76       \\
                                & 1,2        & \textbf{67.34±0.27} & \textbf{68.27±0.65} & \textbf{68.40±0.35}       \\ \hline \bottomrule[1pt]
\end{tabular}
}
\end{table}

\begin{table}[!t]
\centering
    \renewcommand\arraystretch{1.5}
    \caption{Ablation Study on Various Components of MSE-Nets. The Best Results are in Bold.}
    \label{tab3}
    \large
\scalebox{0.8}{

\begin{tabular}{c|c|c|c|c|c}
\toprule[1pt] \hline
Methods                         & $\mathcal{L}_{pc}$ & $\mathcal{L}_{ps}$ & $\mathbf{D}_m$ = 30              & $\mathbf{D}_m$ = 50              & $\mathbf{D}_m$ = 70 \\ \hline \hline
\multirow{4}{*}{MSE-Nets} & $\times$     & $\times$     & 66.00±0.61                    &  66.02±1.55                   & 67.36±1.12       \\
                                & $\times$     & $\checkmark$    & 66.45±0.23                     & 67.31±0.53                    & 67.86±0.57       \\
                                & $\checkmark$    & $\times$     & 66.69±0.72          & 66.53±0.43          & 68.24±0.48       \\
                                & $\checkmark$    & $\checkmark$    & \textbf{67.34±0.27} & \textbf{68.27±0.65} &\textbf{68.40±0.35}        \\ \hline \bottomrule[1pt]
\end{tabular}}
\end{table}
\subsection{Total Loss Function}
Each network consists of three parts of loss, which are multi-annotated agreement loss $\mathcal{L}_{ma}$, prediction consistency loss $\mathcal{L}_{pc}$, and pseudo-supervised loss $\mathcal{L}_{ps}$ for unannotated data. 

The total loss of the $k^{th}$ network is respected by:
\begin{equation}
\begin{split}   
\mathcal{L}_{total}^{k} = \alpha\mathcal{L}_{ma}^k + \beta \mathcal{L}_{pc}^k + \lambda \mathcal{L}_{ps}^k,
\end{split}
\end{equation}
and the total loss is respected by:
\begin{equation}
\begin{split}   
\mathcal{L}_{total} = \sum_{k=1}^{K}{\mathcal{L}_{total}^{k}}.
\end{split}
\end{equation}

Empirically, $\alpha$ and $\beta$ are hyper-parameters and we set $\alpha = 1$, $\beta = 1$. $\lambda$ is a ramp-up trade-off weight commonly scheduled by the time-dependent Gaussian function \cite{cui2019semi} $\lambda(t) = w_{max} \cdot e^{(-5(1-\frac{t}{t_{max}})^2)}$, where $w_{max}$ is the maximum weight commonly set as 0.1 \cite{yu2019uncertainty} and $t_{max}$ is the maximum training iteration. Such a $\lambda$ weight representation avoids being dominated by misleading targets when starting online training.

\subsection{Average Fusion for Predictions}
The prediction results of all network are denoted as $\{\mathbf{P}_1, \mathbf{P}_2,...,\mathbf{P}_K\}$. We use the average fusion to generate the ultimate predicted probability map:
\begin{equation}
\begin{split}
\overline{\mathbf{P}} = \frac{1}{K}\sum_{k=1}^{K}\mathbf{P}_{k}.
\end{split}
\end{equation}
Finally, the final mask is obtained by the predicted probability map $\overline{\mathbf{P}}$.

The benefits of average fusion are as follows:

(i) Increased Robustness: By combining predictions from multiple networks, the final probability map becomes more robust to individual network variations or errors. 

(ii) Mitigation of Overconfidence: Averaging fusion can help in mitigating the issue of overconfidence exhibited by individual networks. 

(iii) Enhanced Generalization: Combining predictions from multiple networks helps to capture a broader range of patterns and variations in the data.

The ablation experiments and effects for average fusion will also be reflected in subsequent chapters.

\begin{table*}[!t]
\centering
    \renewcommand\arraystretch{1.5}
    \caption{Quantitative Results for Different Methods on the RIGA Test Set. The Training and Test Set Is Set as Average Weight Majority Vote, Random Condition, STAPLE strategy~\cite{warfield2004simultaneous}. These Results Are Evaluated Using ($\mathcal{D}_{disc}^s(\%)$, $\mathcal{D}_{cup}^s(\%)$), with the Best Results Indicated in Bold.}
    \label{tab4}
    \large
\scalebox{0.8}{
\begin{tabular}{c|c|cc|cc|cc}
\toprule[1pt] \hline
\multirow{2}{*}{\textbf{Methods}} & \multirow{2}{*}{\textbf{$\mathbf{D}_m$/$\mathbf{D}_u$}} & \multicolumn{2}{c|}{\textbf{Average}}       & \multicolumn{2}{c|}{\textbf{Random}}        & \multicolumn{2}{c}{\textbf{STAPLE}}                              \\ \cline{3-8} 
                         &                                    & \multicolumn{1}{c|}{$\mathcal{D}_{disc}^s(\%)$} & $\mathcal{D}_{cup}^s(\%)$  & \multicolumn{1}{c|}{$\mathcal{D}_{disc}^s(\%)$} & $\mathcal{D}_{cup}^s(\%)$  & \multicolumn{1}{c|}{$\mathcal{D}_{disc}^s(\%)$} & $\mathcal{D}_{cup}^s(\%)$                       \\ \hline \hline
LinkNet                 & 70/0                               & \multicolumn{1}{c|}{91.24} & 86.09 & \multicolumn{1}{c|}{88.36} & 79.82 & \multicolumn{1}{c|}{92.07} &85.09                       \\ \hline \hline
MT~\cite{tarvainen2017mean}                       & 70/585                             & \multicolumn{1}{c|}{92.02} & 86.76 & \multicolumn{1}{c|}{87.77} & 79.02 & \multicolumn{1}{c|}{91.99} & 85.60                      \\ \hline
UAMT~\cite{yu2019uncertainty}                     & 70/585                             & \multicolumn{1}{c|}{91.92} & 86.40 & \multicolumn{1}{c|}{87.75} & 79.01 & \multicolumn{1}{c|}{92.83} & 85.54                      \\ \hline
CPS~\cite{chen2021semi}                      & 70/585                             & \multicolumn{1}{c|}{91.85} & 86.62 & \multicolumn{1}{c|}{87.74} & 79.00 & \multicolumn{1}{c|}{92.47} & \multicolumn{1}{c}{85.20} \\ \hline
ICT~\cite{verma2022interpolation}                      & 70/585                             & \multicolumn{1}{c|}{91.74} & 86.87 & \multicolumn{1}{c|}{87.47} & 78.42 & \multicolumn{1}{c|}{92.59} & \multicolumn{1}{c}{85.51} \\ \hline
BCP~\cite{bai2023bidirectional}                      & 70/585                             & \multicolumn{1}{c|}{91.49} & 86.62 & \multicolumn{1}{c|}{88.23} & 81.40 & \multicolumn{1}{c|}{91.98} & \multicolumn{1}{c}{84.43} \\ \hline \hline
\textbf{MSE-Nets (Ours)}          & \textbf{70/585}                             & \multicolumn{1}{c|}{\textbf{92.38}} & \textbf{87.22} & \multicolumn{1}{c|}{\textbf{89.50}} & \textbf{81.42} & \multicolumn{1}{c|}{\textbf{92.89}} & \multicolumn{1}{c}{\textbf{85.79}} \\ \hline \bottomrule[1pt]
\end{tabular}}
\end{table*}

\section{Experiments}

\subsection{Datasets and Experimental Setup}
We selected the ISIC~\cite{chaurasia2017linknet,codella2018skin} dataset and the RIGA~\cite{almazroa2017agreement} dataset for our experiments.
The ISIC and RIGA dataset consists of images with ambiguous boundaries for the segmentation targets. 
The ISIC dataset provides annotations from two different sources, making it suitable for evaluating the performance of our method under diverse annotation scenarios.
The RIGA dataset offers annotations from six different sources, presenting a more challenging scenario for our method.

\begin{itemize}
\item[$\bullet$] \textbf{ISIC}: We randomly select a subset of images from the multi-annotated ISIC dataset constructed by \cite{ribeiro2020less}. Since part of our work lies in investigating the scarce-data scenario, the training and validation sets contain a total of 250 images and each image contains two ground truth masks, of which the training set contains 200 images and the validation set contains 50 images. We divide the training set into multi-annotated data and unannotated data according to different experimental settings.
To evaluate the segmentation performance of our method, we employ the ISIC test set introduced by~\cite{ribeiro2020less}. The test set comprises a random selection of 2000 images from the ISIC archive, with each image having only one corresponding segmentation ground truth.
\begin{figure}[htb]
\centerline{\includegraphics[width=\linewidth]{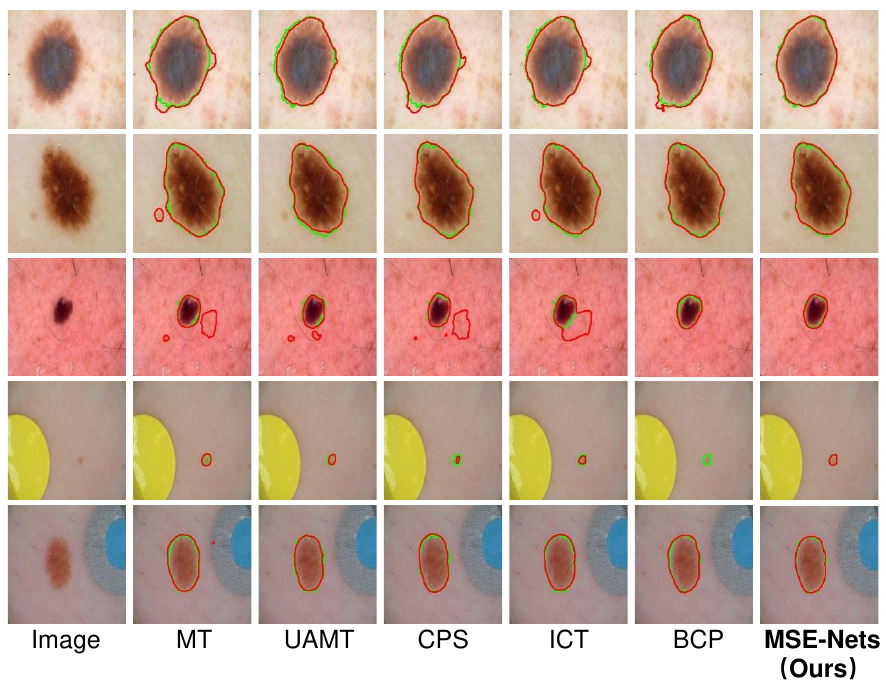}}
    \caption{Visualized segmentation results on ISIC ($\mathbf{D}_m = 50$) of different methods (the results of using effective networks in comparative semi-supervised methods). The ground truth mask and predicted mask are represented by green and red, respectively.}
    \label{fig5}
\end{figure}

\begin{figure}[htb]
\centerline{\includegraphics[width=\linewidth]{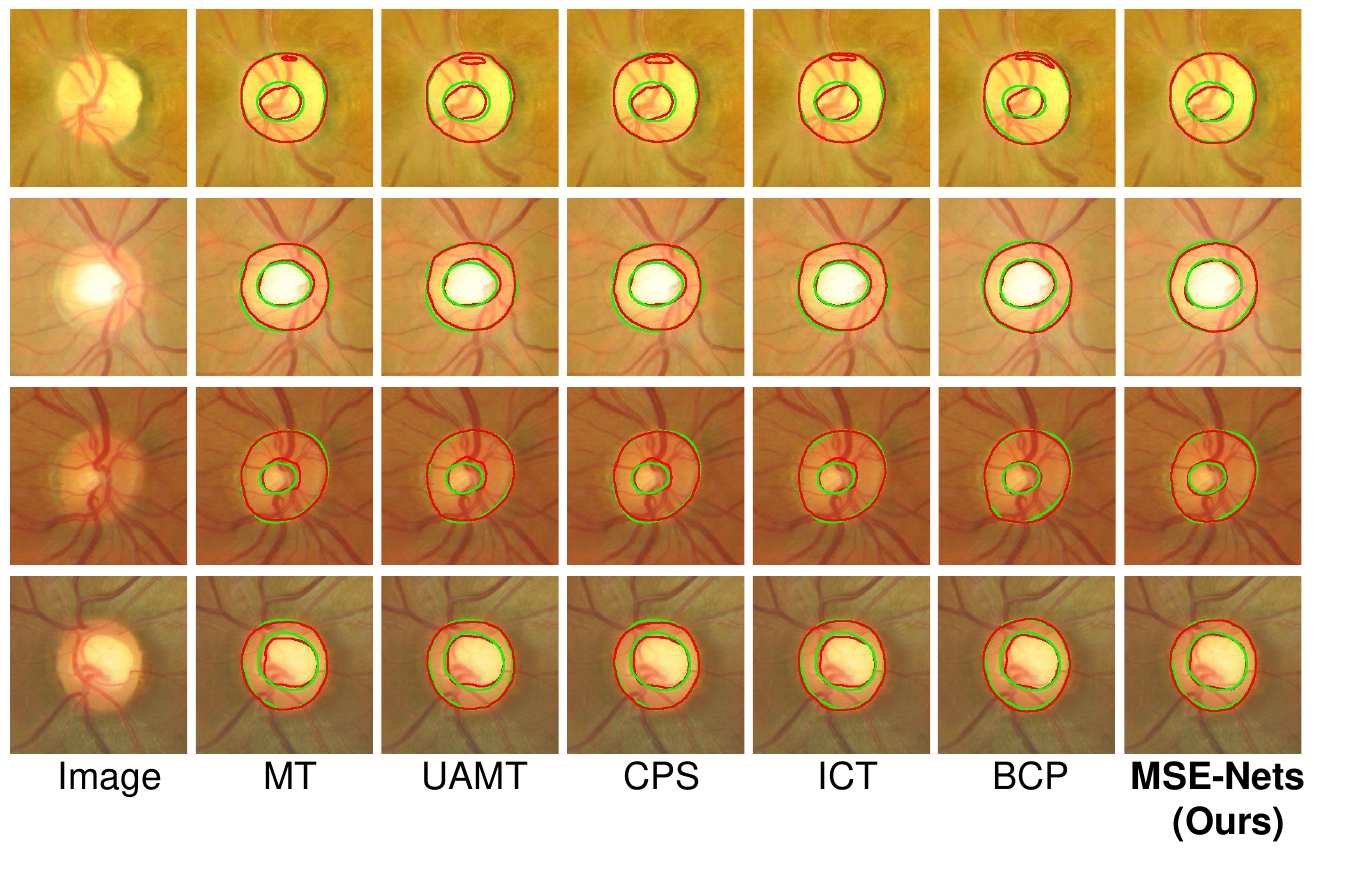}}
    \caption{Visualized segmentation results on RIGA (AVERAGE) of different methods showing optic disc and cup. The ground truth mask and predicted mask are represented by green and red, respectively.}
    \label{fig6}
\end{figure}

\item[$\bullet$] \textbf{RIGA}: The dataset is a publicly available retinal disc and cup segmentation dataset comprising 750 color fundus images from three different sources: 460 images from MESSIDOR, 195 images from BinRushed, and 95 images from Magrabia. The segmentation masks for the optic disc and optic cup outlines were manually annotated by six glaucoma specialists, following the RIGA benchmark~\cite{almazroa2017agreement}. 
During model training, we select 195 samples from BinRushed and 460 samples from MESSIDOR as training sets, as follows~\cite{ji2021learning}. We randomly selected 70 samples as multi-annotated data, and the rest of the images as unannotated data.
The Magrabia set with 95 samples is chosen as the test set to evaluate the model. The total training and validation set both contain six segmentation ground truth masks. It can be directly used as input data and evaluation data for our method. Considering the limitations of comparative semi-supervised methods on multi-annotated data, we use three different methods to construct three training (test) sets: Average Weight Majority Vote (Average), Random selection (Random), and STAPLE strategy~\cite{warfield2004simultaneous} (STAPLE). By using the RIGA dataset, we aimed to investigate the effectiveness of our approach in handling more annotations with diverse characteristics.
\end{itemize}

\begin{table*}[!t]
\centering
    \renewcommand\arraystretch{1.5}
    \caption{The Ablation Study of MSE-Nets on RIGA Test Set. The Table Shows the results Obtained by Adopting the Average Fusion Strategy during Inference (the model that works best on every single annotation). These Results Are Evaluated Using ($\mathcal{D}_{disc}^s(\%)$, $\mathcal{D}_{cup}^s(\%)$), with the Best Results Indicated in Bold.}
    \label{tab5}
    \large
\scalebox{0.75}{

\begin{tabular}{c|c|cc|cc|cc}
\toprule[1pt] \hline
\multirow{2}{*}{\textbf{Methods}} & \multirow{2}{*}{\textbf{Annotators}} & \multicolumn{2}{c|}{\textbf{Average}}       & \multicolumn{2}{c|}{\textbf{Random}}        & \multicolumn{2}{c}{\textbf{STAPLE}}         \\ \cline{3-8} 
                         &                             & \multicolumn{1}{c|}{$\mathcal{D}_{disc}^s(\%)$} & $\mathcal{D}_{cup}^s(\%)$  & \multicolumn{1}{c|}{$\mathcal{D}_{disc}^s(\%)$} & $\mathcal{D}_{cup}^s(\%)$  & \multicolumn{1}{c|}{$\mathcal{D}_{disc}^s(\%)$} & $\mathcal{D}_{cup}^s(\%)$  \\ \hline \hline
LinkNet          & 1-6                         & \multicolumn{1}{c|}{91.78} & 86.65 & \multicolumn{1}{c|}{88.98} & 80.82 & \multicolumn{1}{c|}{92.38} & 85.24 \\ \hline \hline
MT~\cite{tarvainen2017mean}                       & 1-6                         & \multicolumn{1}{c|}{92.15} & 87.13 & \multicolumn{1}{c|}{89.30} & 81.34 & \multicolumn{1}{c|}{92.64} & 85.42 \\ \hline
UAMT~\cite{yu2019uncertainty}                     & 1-6                         & \multicolumn{1}{c|}{92.11} & 87.02 & \multicolumn{1}{c|}{89.27} & 81.37 & \multicolumn{1}{c|}{92.64} & 85.56 \\ \hline
CPS~\cite{chen2021semi}                      & 1-6                         & \multicolumn{1}{c|}{92.07} & 86.80 & \multicolumn{1}{c|}{89.20} & 81.09 & \multicolumn{1}{c|}{92.58} & 85.27 \\ \hline
ICT~\cite{verma2022interpolation}                      & 1-6                         & \multicolumn{1}{c|}{91.45} & 85.46 & \multicolumn{1}{c|}{89.12} & 81.25 & \multicolumn{1}{c|}{92.50} & 85.31 \\ \hline
BCP~\cite{bai2023bidirectional}                      & 1-6                         & \multicolumn{1}{c|}{91.46} & 86.66 & \multicolumn{1}{c|}{88.78} & 80.86 & \multicolumn{1}{c|}{92.12} & 85.47           \\ \hline \hline
\textbf{MSE-Nets (Ours)}          & \textbf{1-6}                         & \multicolumn{1}{c|}{\textbf{92.38}} & \textbf{87.22}& \multicolumn{1}{c|}{\textbf{89.50}} & \textbf{81.42} & \multicolumn{1}{c|}{\textbf{92.89}} & \textbf{85.79} \\ \hline \bottomrule[1pt]
\end{tabular}}
\end{table*}

\subsubsection{Baseline Approaches}
We seek to include as many different baselines as possible, providing insights for future research. Specifically, baselines can be divided into the following categories:
\begin{itemize}
\item[$\bullet$] Fully-supervised baselines: $\textbf{LinkNet}$~\cite{chaurasia2017linknet}: the backbone trained using only annotated data.
\end{itemize}

\begin{itemize}
\item[$\bullet$] Semi-supervised baselines: 
$\textbf{MT}$~\cite{tarvainen2017mean}: encourages prediction consistency between the student model and the teacher model.
$\textbf{UAMT}$~\cite{yu2019uncertainty}: utilizes uncertainty information, the student network is guided to progressively learn from valuable and dependable targets provided by the teacher network.
$\textbf{CPS}$~\cite{chen2021semi}: uses two networks with the same structure but different initializations, adding constraints to ensure that the output of both networks for the same sample exhibits similarity.
$\textbf{ICT}$~\cite{verma2022interpolation}: encourages the coherence between the prediction at an interpolation of unlabeled points and the interpolation of the predictions at those points.
$\textbf{BCP}$~\cite{bai2023bidirectional}: introduces a bidirectional CutMix approach to facilitate comprehensive learning of common semantics from labeled and unlabeled data in both inward and outward directions.
\end{itemize}

Moreover, in order to further compare the performance of our method, we introduce multi-annotation methods LIS~\cite{ribeiro2020less} and D-LEMA~\cite{mirikharaji2021d} in the fully-supervised comparison of ISIC.

\subsubsection{Implementation and Evaluation Metrics}
We implemented our method and in Python using PyTorch and performed the computations on an NVIDIA GeForce RTX 3090 GPU with 24GB of memory. All networks use different initializations and are trained using the Adam optimizer (betas=(0.9, 0.99)). The initial learning rate is set to 1e-4 and is divided by 10 every 2000 iterations. Images fed into the network are resized to 256 × 256 pixels and normalized using per-channel mean and standard deviation. The model with the best performance (each network in the corresponding validation set) on the validation set is selected as the final model.
\begin{itemize}
\item[$\bullet$] \textbf{ISIC}: The batch size is set to 4 and the multi-annotated image for each iteration (total of 15000 iterations) is 1.  We utilize the recognized metrics Jaccard index and repeat the experiment five times to report the mean and standard error.
\item[$\bullet$] \textbf{RIGA}: The batch size is set to 8 and the multi-annotated image for each iteration (total of 40000 iterations) is 1. We present the Dice Similarity Coefficient (DSC) for each category, excluding the background.
\end{itemize}

\subsection{Experiments on ISIC Dataset}

\subsubsection{Comparison Study}
Table~\ref{tab1} shows the results of different benchmark methods and MSE-Nets on the different $\mathbf{D}_m$ of the ISIC. The first section of the table displays the results obtained by the LinkNet architecture under two different annotators. We can observe that LinkNet achieves moderate segmentation accuracy with some variation in performance across different annotators. When we turn to semi-supervised methods, MT and UAMT demonstrate similar performance, yielding a moderate Jaccard index. These semi-supervised methods provide acceptable segmentation results but may encounter challenges in accurately capturing boundaries. Furthermore, the performance of ICT, CPS, and BCP is comparable to the above methods, indicating that they are also effective in generating segmentation results. It is worth noting that the performance of some semi-supervised (ICT, BCP) decreases when additional annotation data is introduced. This can be attributed to the introduction of additional misinformation in the annotations, particularly in cases where the boundaries of the segmented objects are ambiguous. The presence of ambiguous boundaries makes it challenging to accurately define the boundaries, leading to a decrease in segmentation performance.

Although part of semi-supervised models trained with a single annotator can also improve accuracy, our method achieves the best results, with a Jaccard index improvement of around 3\% in the setting of $\mathbf{D}_m=30$ and $\mathbf{D}_m=50$ compared to the single-annotator approach. Furthermore, this improvement extends to around 4\% in the case of $\mathbf{D}_m=70$. The smaller standard error further attests to the robustness of our method.
Moreover, compared to the fully-supervised methods LIS and D-LEMA with $\mathbf{D}_m = 2333$, our method has nearly approached the performance of the LIS, considering that our multi-annotated data is only about 3\% of its volume. The difference from D-LEMA is also within an acceptable range.
This confirms that our method can effectively integrate different ground truth masks to find the correct mask, thereby eliminating noise between different. 
The results highlight the effectiveness of the MSE-Nets method in enhancing the segmentation of medical images, surpassing the performance of the other evaluated methods. 
The visualization results of different methods on the test set are presented in Fig.~\ref{fig5}, further confirming the effectiveness of our approach in addressing semi-supervised learning with ambiguous boundaries. 
In the fourth row of visualization results, the method of BCP did not identify the lesion area, while the lesion area identified by our method was almost the same as the ground truth.
These visualizations highlight the superior performance of our method in improving image segmentation with ambiguous boundaries, reinforcing its capabilities in this challenging domain.
\begin{table*}[!t]
\centering
    \renewcommand\arraystretch{1.5}
    \caption{The Ablation Study Of MSE-Nets on RIGA Test Set. The Table Shows the Results of Baseline and MSE-Nets on Different $K$ (2-6) multi-annotated data. These Results Are Evaluated Using ($\mathcal{D}_{disc}^s(\%)$, $\mathcal{D}_{cup}^s(\%)$), with the Best Results Indicated in Bold.}
    \label{tab6}
    \large
\scalebox{0.8}{
\begin{tabular}{c|c|cc|cc|cc}
\toprule[1pt] \hline
\multirow{2}{*}{\textbf{Methods}}         & \multirow{2}{*}{\textbf{Annotators}} & \multicolumn{2}{c|}{\textbf{Average}}       & \multicolumn{2}{c|}{\textbf{Random}}        & \multicolumn{2}{c}{\textbf{STAPLE}}         \\ \cline{3-8} 
                                 &                             & \multicolumn{1}{c|}{$\mathcal{D}_{disc}^s(\%)$} & $\mathcal{D}_{cup}^s(\%)$  & \multicolumn{1}{c|}{$\mathcal{D}_{disc}^s(\%)$} & $\mathcal{D}_{cup}^s(\%)$  & \multicolumn{1}{c|}{$\mathcal{D}_{disc}^s(\%)$} & $\mathcal{D}_{cup}^s(\%)$  \\ \hline \hline
LinkNet                         & 1                           & \multicolumn{1}{c|}{90.32} & 83.64 & \multicolumn{1}{c|}{88.13} & 78.43 & \multicolumn{1}{c|}{91.39} & 83.38 \\ \hline \hline
\multirow{5}{*}{MSE-Nets} & 1-2                         & \multicolumn{1}{c|}{91.41} & 86.23 & \multicolumn{1}{c|}{88.86} & 80.66 & \multicolumn{1}{c|}{92.15} & 85.06 \\ \cline{2-8} 
                                 & 1-3                         & \multicolumn{1}{c|}{91.87} & 86.16 & \multicolumn{1}{c|}{89.14} & 80.53 & \multicolumn{1}{c|}{92.64} & 85.26 \\ \cline{2-8} 
                                 & 1-4                         & \multicolumn{1}{c|}{91.99} & 87.04 & \multicolumn{1}{c|}{89.09} & 80.96 & \multicolumn{1}{c|}{92.24} & 84.49 \\ \cline{2-8} 
                                 & 1-5                         & \multicolumn{1}{c|}{92.26} & 87.18 & \multicolumn{1}{c|}{89.25} & 81.10 & \multicolumn{1}{c|}{92.62} & 85.21 \\ \cline{2-8} 
                                 & 1-6                         & \multicolumn{1}{c|}{\textbf{92.38}} & \textbf{87.22} & \multicolumn{1}{c|}{\textbf{89.50}} & \textbf{81.42} & \multicolumn{1}{c|}{\textbf{92.89}} & \textbf{85.79} \\ \hline \bottomrule[1pt]
\end{tabular}}
\end{table*}
\begin{figure*}[!t]
\centerline{\includegraphics[width=\linewidth]{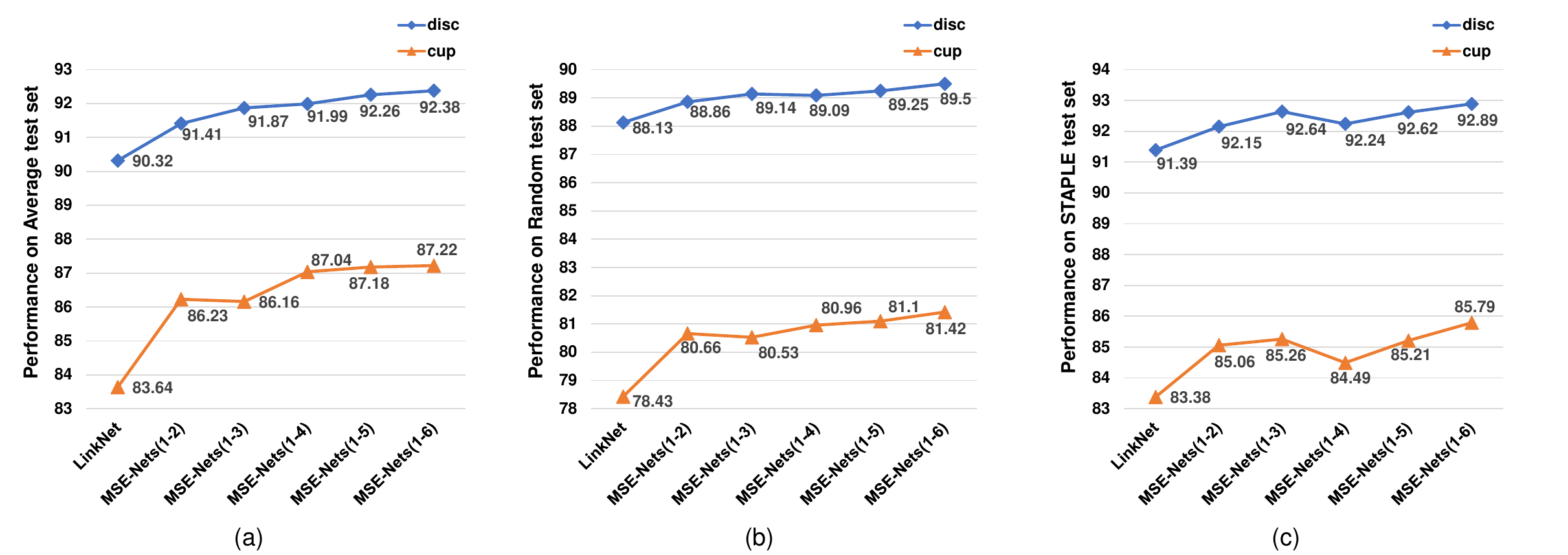}}
    \caption{Visualisation of line graph results for different number of annotations obtained for baseline and MSE-Nets. (a) Performance of Average test set with baseline and different $K$ of MSE-Nets. (b) Performance of Random test set with baseline and different $K$ of MSE-Nets. (c) Performance of STAPLE test set with baseline and different $K$ of MSE-Nets.}
    \label{fig7}
\end{figure*}

\subsubsection{Analytical Ablation Study}
To evaluate the effectiveness of each component of MSE-Nets on the ISIC dataset, we have conducted ablation studies using different variants. Our ablation experiments are presented in Table~\ref{tab2}, and Table~\ref{tab3}, showcasing the results obtained.

Table~\ref{tab2} shows the difference in the experimental results of whether the fusion strategy is used during inference. We can obtain better experimental results by using average fusion during inference compared to any single individual network. There are several benefits of employing average fusion during inference.
Firstly, average fusion allows models to capture the diversity and complementary information from different networks by integrating prediction results. This integration enhances the robustness and accuracy of the segmentation results.
Secondly, average fusion helps mitigate the impact of noise and uncertainty present in individual prediction results. By combining multiple predictions, we can reduce or even eliminate errors and inconsistencies introduced by individual networks, thereby improving the overall segmentation quality.
Additionally, average fusion leverages the strengths and expertise of different networks. Each predicted mask from the network may have biases, limitations, or specialized knowledge in certain aspects of the segmentation task. By fusing their prediction results, we can harness their unique abilities to achieve more comprehensive and refined segmentation outputs.
In summary, employing average fusion during inference allows us to leverage the collective wisdom and knowledge of multiple prediction results, leading to improved segmentation performance and increased confidence in the results.

Table~\ref{tab3} shows the impact of the different losses on the results of our proposed method on the ISIC dataset. The absence of $\mathcal{L}_{ps}$ and the lack of $\mathcal{L}_{pc}$ (only use pixels with agreement annotations for training) give relatively poor performance of MSE-Nets.
However, for other semi-supervised methods, the results are much better, demonstrating that pixel-level pairwise agreement separation can yield reliable labels.
Furthermore, Training the MSE-Nets with either $\mathcal{L}_{ps}$ or $\mathcal{L}_{pc}$ as a consistency constraint yields improved results, which also indicates that consistent pseudo-labels provide a performance boost to the model.
The results obtained by combining $\mathcal{L}_{pc}$, and $\mathcal{L}_{ps}$ training achieved the highest segmentation accuracy compared to other methods. The results of the ablation study suggest that incorporating all losses leads to the most effective MSE-Nets model for semi-supervised medical image segmentation with ambiguous boundaries. The advantage of MSE-Nets is that it can capture richer boundary information. These findings demonstrate the effectiveness of incorporating both annotation agreement constraints and prediction consistency to enhance the performance of our method. By leveraging multiple sources of consistency constraints, our method achieves exceptional segmentation accuracy.

\subsection{Experiments on RIGA}
\subsubsection{Comparison Study}
Table~\ref{tab4} shows the results of different benchmark methods and MSE-Nets on the RIGA dataset.  Fully-supervised networks (LinkNet) trained using only a subset of annotated data did not achieve satisfactory results, with most of the semi-supervised networks surpassing the fully-supervised baselines on both the optic disc and cup segmentation tasks. Compared to other semi-supervised methods, including MT, UAMT, CPS, ICT, and BCP, MSE-Nets achieved the best results for the metrics on three different scenarios (Average, Random, STAPLE). Our approach significantly outperforms other semi-supervised methods Particularly, under the Average and STAPLE strategies, the MSE-Nets achieves the $\mathcal{D}_{disc}^s$ and $\mathcal{D}_{cup}^s$ values, reaching 92.38\% and 87.22\%, 92.89\% and 85.79\%, respectively. 

In Fig.~\ref{fig6}, we can observe the visualization results of different methods, further confirming our proposed method's effectiveness. The visualizations showcase the ability of our approach to accurately segment the desired regions, aligning with the corresponding ground truth annotations. These visual results provide qualitative evidence of the superior performance of our method compared to the other methods.

\subsubsection{Analytical Ablation Study}
To further demonstrate that the gains of our method are not simply the result of average fusing, we train (evaluate) the network on individual annotator's training (testing) sets. After obtaining all the best-performing networks, we perform an average fusion to derive the final result.

Table~\ref{tab5} presents the comparison results between MSE-Nets and all the other semi-supervised methods using average fusion during inference. By utilizing multiple networks and averaging their predictions, we introduce an ensemble effect that aids in generalization and reduces the risk of over-fitting. Other methods that use single-annotation training and average fusion of probability prediction maps at inference time cannot effectively remove noise and bias. However, our method adopts multi-annotated training so that each network can learn reliable information about all annotations. From the table, we can see that when the inference is average fusion, MSE-Nets outperforms the other methods on the metrics when inference is averaged over fusion.

To further investigate the performance of MSE-Nets on more diverse expert annotations, Table~\ref{tab6} presents the results of MSE-Nets ablation studies on multiple annotators with different numbers (from 2 to 6). 

As we observed, the performance improves significantly as the number of annotators increases. When utilizing annotations from 1-2 or 1-3 annotators, the dice scores are already relatively high, showing that even a small number of multi-annotators can improve segmentation accuracy. When we consider annotations from 1-4 and 1-5 annotators, the dice scores further improve, which demonstrates that our method can incorporate more diverse viewpoints from multiple annotators to enhance the segmentation ability of the network. Interestingly, when considering annotations from all six annotators (1-6), the performance reaches its highest point. Fig.~\ref{fig7} demonstrates the performance of the baseline and MSE-Nets on the Average, Random, and STAPLE test sets. We observe that (1) the MSE-Nets obtained for different annotation quantities $K$ outperform the baseline, and (2) the performance of MSE-Nets gradually increases with the increase in $K$.
The result confirms that utilizing the consensus of multiple annotators can lead to more comprehensive and accurate segmentation models. Our method can combine different annotations from multiple annotators to cope with ambiguous boundary scenarios in semi-supervised medical image segmentation tasks.
Overall, the ablation study underscores the effectiveness of MSE-Nets in exploiting multi-annotated data and capitalizing on the diversity among annotators to achieve superior segmentation results. By considering multiple viewpoints, we can reduce bias and errors individual annotators make at ambiguous boundaries.

\begin{figure}[!t]
\centerline{\includegraphics[width=\linewidth]{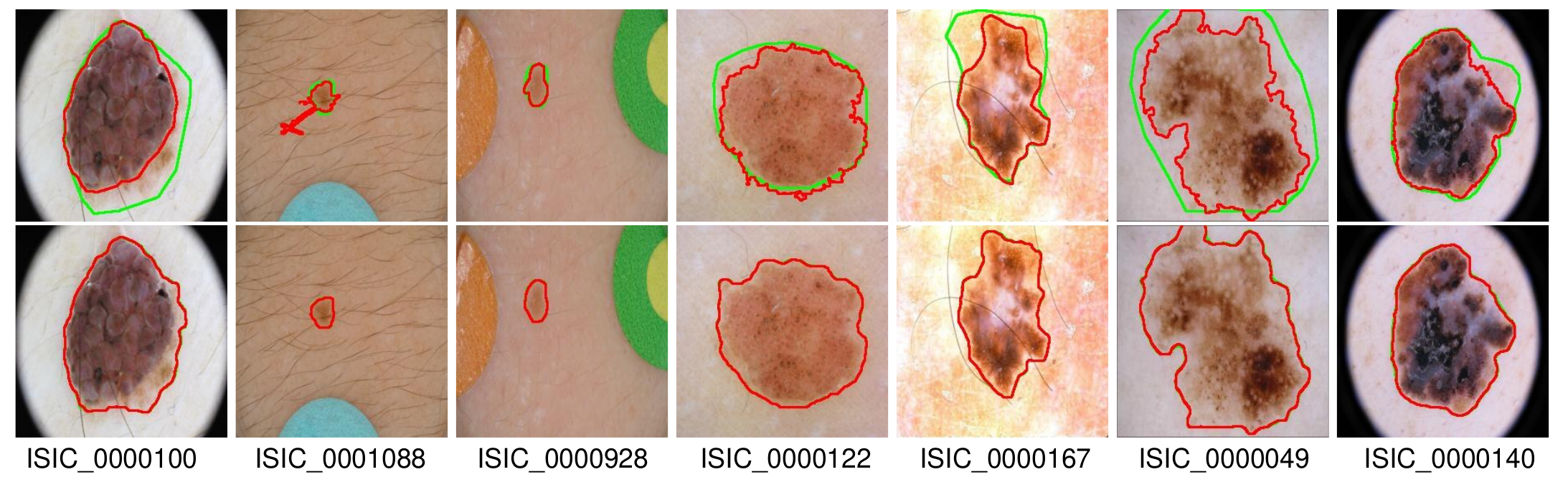}}
    \caption{Visualized segmentation results of different annotations on the ISIC ($\mathbf{D}_m = 50$) dataset at the beginning and end of training. The top images represent the annotations at the start of training, while the bottom images represent the annotations after training completion (in different colors).}
    \label{fig8}
\end{figure}

\section{Discussion}
Although the use of multiple annotations per image has been extensively studied in the fully-supervised setting, obtaining large amounts of multi-annotated data is challenging due to the significant time and human cost required to segment annotations, so most images lack any annotations.
In this study, we propose MSE-Nets, and the proposed NPCE is based on a pixel-level label selection strategy and combined with a label refinement strategy to fully utilize multi-annotated data. Furthermore, the proposed MNPS is based on the consistent pseudo-label strategy to avoid imprecise pseudo-label of unannotated data from negatively affecting the network.

To further illustrate the effectiveness of NPCE, Fig.~\ref{fig8} shows the comparison of different annotations on ISIC training set images at the beginning and end of training. The visual annotation displayed at the end of training appears to have only one color due to the high similarity between the two masks. This shows a clear trend of increasing consistency among initially different annotations as training proceeds. The visualization results demonstrate that the NPCE module avoids favoring any specific annotation and instead guides them toward more accurate representations. The visual results further confirm the effectiveness of our approach in refining the annotations and improving the accuracy of boundary delineation. The transformation from initially uncertain boundary annotations to more consistent and accurate representations showcases the potential of our method in handling the challenges associated with ambiguous boundaries in medical image segmentation. Our findings demonstrate the potential of leveraging multi-annotated information and exploiting diversity among annotators to achieve better segmentation results. The ability to learn from multiple annotations and guide them toward a more cohesive consensus significantly enhances the robustness and generalizability of the segmentation model.

Our proposed method has demonstrated encouraging results, however, there are still areas that warrant improvement. One area of concern is the empirical distribution mismatch~\cite{bai2023bidirectional} between multi-annotated data and unannotated data, which we did not explicitly address in this work. When treating multi-annotated and unannotated data separately or inconsistently, the knowledge learned from multi-annotated data might be underutilized, leading to suboptimal segmentation performance and increased training time. To mitigate this issue, we plan to explore data augmentation, such as CutMix, to reduce the impact of distribution differences and enhance the network's ability to generalize to unseen data.

Additionally, our current framework only considers the agreement (consistency) of two annotators (network predictions) for multi-annotated to guide the learning process. In future research, we intend to investigate the benefits of incorporating multiple annotators or achieving a more consistent consensus of network predictions. Leveraging a broader range of annotations could potentially lead to more accurate segmentation results.

As semi-supervised medical image segmentation continues to evolve, we aim to explore other advanced methods that can handle ambiguous boundaries more effectively. This may involve incorporating domain knowledge or leveraging advanced deep-learning architectures that are specifically designed to address the challenges posed by uncertain and ambiguous boundaries in medical images.

\section{Conclusion}
In this study, we propose Multi-annotated Semi-supervised Ensemble Networks (MSE-Nets) for learning medical image segmentation with ambiguous boundaries.
In the context of multi-annotated semi-supervised scenarios, characterized by the substantial time and cost implications associated with manual annotations, datasets frequently consist of a limited quantity of multi-annotated data alongside a substantial volume of unannotated data.
To address this, we propose two different modules NPCE and MNPS to handle multi-annotated data and unannotated data respectively.
The proposed NPCE module can make full use of annotation information at the pixel level by comparing annotations between different experts and refining unreliable annotations by network predictions.
As for the majority of the unannotated data, the proposed MNPS module takes the consistent mask of multiple network predictions as a basic fact, thus avoiding the detrimental effects of imprecise pseudo-labels on network learning.
Through extensive experiments on the ISIC and RIGA datasets, our proposed method performs well in semi-supervised segmentation tasks with ambiguous boundaries, compared with other semi-supervised methods that only use a single annotation or a combined fusion approach.
Furthermore, our method excels in capturing object boundaries and generating prediction masks. The visualization results serve as a testament to the outstanding performance of our proposed approach.

\bibliographystyle{IEEEtran}
\bibliography{IEEEabrv,myrefs}

\end{document}